\documentclass[dvipsnames,format=sigconf,anonymous=false,review=false]{acmart}
\AtBeginDocument{%
  }

\setcopyright{acmlicensed}
\acmConference[GECCO '25]{The Genetic and Evolutionary Computation Conference 2025}{July 14--18, 2025}{Málaga, Spain}

\usepackage[subtle]{savetrees}

\begin{document}

\title{Lifelong Evolution of Swarms}

\author{Lorenzo Leuzzi}
\email{l.leuzzi1@studenti.unipi.it}
\affiliation{%
  \institution{University of Pisa}
  \city{Pisa}
  \country{Italy}
}

\author{Simon Jones}
\email{simon2.jones@bristol.ac.uk}
\affiliation{%
  \institution{University of Bristol}
  \city{Bristol}
  \country{UK}
}

\author{Sabine Hauert}
\email{sabine.hauert@bristol.ac.uk}
\affiliation{%
  \institution{University of Bristol}
  \city{Bristol}
  \country{UK}
}

\author{Davide Bacciu}
\email{davide.bacciu@unipi.it}
\affiliation{%
  \institution{University of Pisa}
  \city{Pisa}
  \country{Italy}}

\author{Andrea Cossu}
\email{andrea.cossu@unipi.it}
\affiliation{%
  \institution{University of Pisa}
  \city{Pisa}
  \country{Italy}
}
\renewcommand{\shortauthors}{Leuzzi et al.}

\begin{abstract}
Adapting to task changes without forgetting previous knowledge is a key skill for intelligent systems, and a crucial aspect of lifelong learning. Swarm controllers, however, are typically designed for specific tasks, lacking the ability to retain knowledge across changing tasks. Lifelong learning, on the other hand, focuses on individual agents with limited insights into the emergent abilities of a collective like a swarm.
To address this gap, we introduce a lifelong evolutionary framework for swarms, where a population of swarm controllers is evolved in a dynamic environment that incrementally presents novel tasks. This requires evolution to find controllers that quickly adapt to new tasks while retaining knowledge of previous ones, as they may reappear in the future.
We discover that the population inherently preserves information about previous tasks, and it can reuse it to foster adaptation and mitigate forgetting. In contrast, the top-performing individual for a given task catastrophically forgets previous tasks. To mitigate this phenomenon, we design a regularization process for the evolutionary algorithm, reducing forgetting in top-performing individuals.
Evolving swarms in a lifelong fashion raises fundamental questions on the current state of deep lifelong learning and on the robustness of swarm controllers in dynamic environments.
\end{abstract}



\maketitle
\section{Introduction}
A swarm is composed of simple agents that interact locally and give rise to a decentralized, self-organized behavior towards a common goal. Designing swarm controllers is a critical challenge, as manually defining the individual rules to achieve a desired collective behavior is often unfeasible due to the complex interactions and emergent properties of swarms.
Evolutionary algorithms provide an effective solution to this challenge, bypassing the need for manual intervention and letting the environment decide which solutions are worth exploring more than others. \\
Yet, to truly operate in complex dynamic environments, swarm systems must not only adapt to novel challenges but also retain previously acquired skills. For example, as we will motivate in this paper, a swarm that needs to solve a foraging task may be required to fetch different types of objects and/or bring them to different locations. Changes in the task (e.g., fetching a different object) require the swarm controller to adapt. Subsequent changes in the task should not erase all the information, since previous tasks might occur again. A swarm able to preserve previous knowledge will be able to adapt faster to reoccurring tasks than a swarm that always starts from scratch.\\
This is precisely the focus of \emph{lifelong learning} \cite{parisi2019}, a research topic mainly explored within deep learning \cite{masana2023} and, to a lesser extent, robotics \cite{lesort2020a}. In lifelong learning a single agent, usually an artificial neural network, is trained via backpropagation on a sequence of tasks. The network has to adapt to the new task and preserve the performance on previously encountered tasks.\\
In this paper, we merge the evolutionary and lifelong paradigms and propose a lifelong evolutionary environment (Section \ref{sec:environment}) for the design of swarm controllers (Section \ref{sec:controller}). Our objective is to enable swarms to quickly adapt to new tasks while preserving previous knowledge. We study the performance of the evolved population over the course of several hundred generations, with particular attention to the drifting points when a new task replaces the current one in the environment (Section \ref{sec:quickadapt}). Inspired by lifelong learning, we also study the performance of the \emph{top-performing individual} on a given task, and how much knowledge of previous tasks it retains. This is useful when deploying a single evolved controller in a dynamic environment that can change at any moment. We discovered that, while the population inherently exhibits a certain degree of retention due to its diversity (Section \ref{sec:forgettingpop}), the individual catastrophically forgets previous knowledge (Section \ref{sec:forgettingind}). Similar to regularization approaches in lifelong learning \cite{parisi2019, kirkpatrick2017}, we propose a regularized version of the evolution process that is able to mitigate forgetting for the top-performing individual, achieving a controller that trades-off the performance on previous tasks and adaptation to the current task. \\ The deep lifelong learning perspective based on a single agent may not be the only choice to build adaptive systems that retain knowledge, as we discovered that an evolved population with enough diversity is able to partially satisfy this requirement without an ad-hoc design. On the other hand, ideas from lifelong learning can help guide the evolutionary process, opening new challenges for the design of swarm controllers in dynamic environments.

\section{Background}
The evolution of swarm controllers focuses on environments where the task to solve do not change over time. The literature is currently lacking a thorough study on how evolutionary algorithms behave when the population of controllers needs not only to adapt to a given task but also to retain information from previously encountered tasks. This challenge is instead the focus of lifelong learning, and it is usually studied for single agents implemented by deep artificial neural networks. \\ 
We therefore provide a short overview on lifelong learning first, and on evolution of swarm controllers next. We devote particular attention to those aspects that will be leveraged for our lifelong evolutionary swarms framework.

\paragraph{Lifelong learning.} Lifelong learning \cite{parisi2019}, also called continual learning \cite{cossu2021a}, designs agents that learn from a stream of observations without forgetting previous knowledge. Lifelong learning is mainly studied in the context of deep learning \cite{verwimp2024} and reinforcement learning \cite{khetarpal2022}, where the learning agent is a deep artificial neural network trained with backpropagation. However, its scope is much broader, as it encompasses any environment where the task to solve is drifting over time \cite{giannini2024} and it is not tied to the specific learning approach or model. \\
Recently, lifelong learning research focused on the forgetting issue that plagues neural networks and other predictive models \cite{french1999}: upon learning new information, the model's performance on previously observed data quickly deteriorates (e.g., incrementally adding new classes in a classification problem reduces the model's performance on previously seen classes). Forgetting is mainly due to the network's inability to maintain a stability-plasticity trade-off \cite{carpenter1986}. Overly plastic networks tend to forget previous knowledge, while overly stable networks are unable to adapt to new tasks. To mitigate forgetting, a plethora of different solutions have been developed \cite{masana2023}. One of the most popular approaches to combat forgetting is Elastic Weight Consolidation (EWC) \cite{kirkpatrick2017} and variants thereof \cite{zenke2017, chaudhry2018}. The idea behind these \emph{importance-based regularization approaches} \cite{parisi2019} is to limit the plasticity of synaptic connections that are deemed important for previous tasks. The plasticity is computed with respect to a previous version of the same model that performed well on previous tasks. As a consequence, the network will be forced to leverage the remaining connections for the adaptation to new incoming tasks, with an improved stability of previous performance. Computing the importance of a given synaptic connection is at the heart of importance-based regularization strategies. Often times, the resulting performance represents a trade-off between the ideal performance on new tasks and the ideal performance on previous tasks. We exploit this idea and adapt it in our lifelong evolutionary swarms framework to mitigate forgetting for the top-performing individual of a population.

\paragraph{Swarm robotics} Swarm robotics takes inspiration from swarms in nature with collective behaviour emerging through local interactions between agents and with the environment \cite{csahin2008special}. A fundamental issue is the design of controllers such that a desired collective behaviour emerges, common approaches include bioinspiration, evolution, hand design, reverse engineering \cite{reynolds1987flocks,hauert2009reverse,winfield2009towards, francesca2016automatic,jones2019onboard}. Making swarms adaptive to change is related to lifelong learning \cite{yao2014improving,castello2016adaptive}. See \cite{bredeche2018embodied,birattari2019automatic,schranz2020swarm,dorigo2021swarm} for recent reviews. Much work has focussed on neuroevolution of controllers; one state-of-the-art approach is NeuroEvolution of Augmenting Topologies (NEAT) \cite{Stanley2002}, which we use in our work. NEAT not only searches for optimal neural network weights but also evolves network topologies in search of the best \emph{minimal} architecture. 
NEAT has been effectively utilized to solve highly complex problems, such as double pole balancing, where it outperforms several methods that rely on fixed topologies \cite{Stanley2004}. The algorithm's superior performance can be attributed to three key features: the use of historical markers to enable meaningful crossover between different topologies, a niching mechanism (speciation), and the gradual evolution of topologies starting from simple initial structures (complexification).\\
In swarm robotics, NEAT has been applied to evolve neural networks for autonomous foraging tasks, as demonstrated by NeatFA \cite{Ericksen2017}. This approach achieved performance comparable to or surpassing established algorithms such as the Central Place Foraging Algorithm \cite{Joshua2015} (CPFA) and the Distributed Deterministic Spiral Algorithm (DDSA) \cite{Fricke2016}.
NEAT has also been used in distributed online learning with swarms (odNEAT) \cite{Silva2015}. odNEAT operates across multiple robots which have to solve the same task, either individually or collectively. A significant aspect of odNEAT is its dynamic population management, where each individual maintains an internal set of genomes, including current and previously successful controllers. odNEAT provides results comparable to centralised approaches \cite{Stanley2005}.
\begin{figure}[t]
  \centering
    \includegraphics[width=0.45\textwidth]{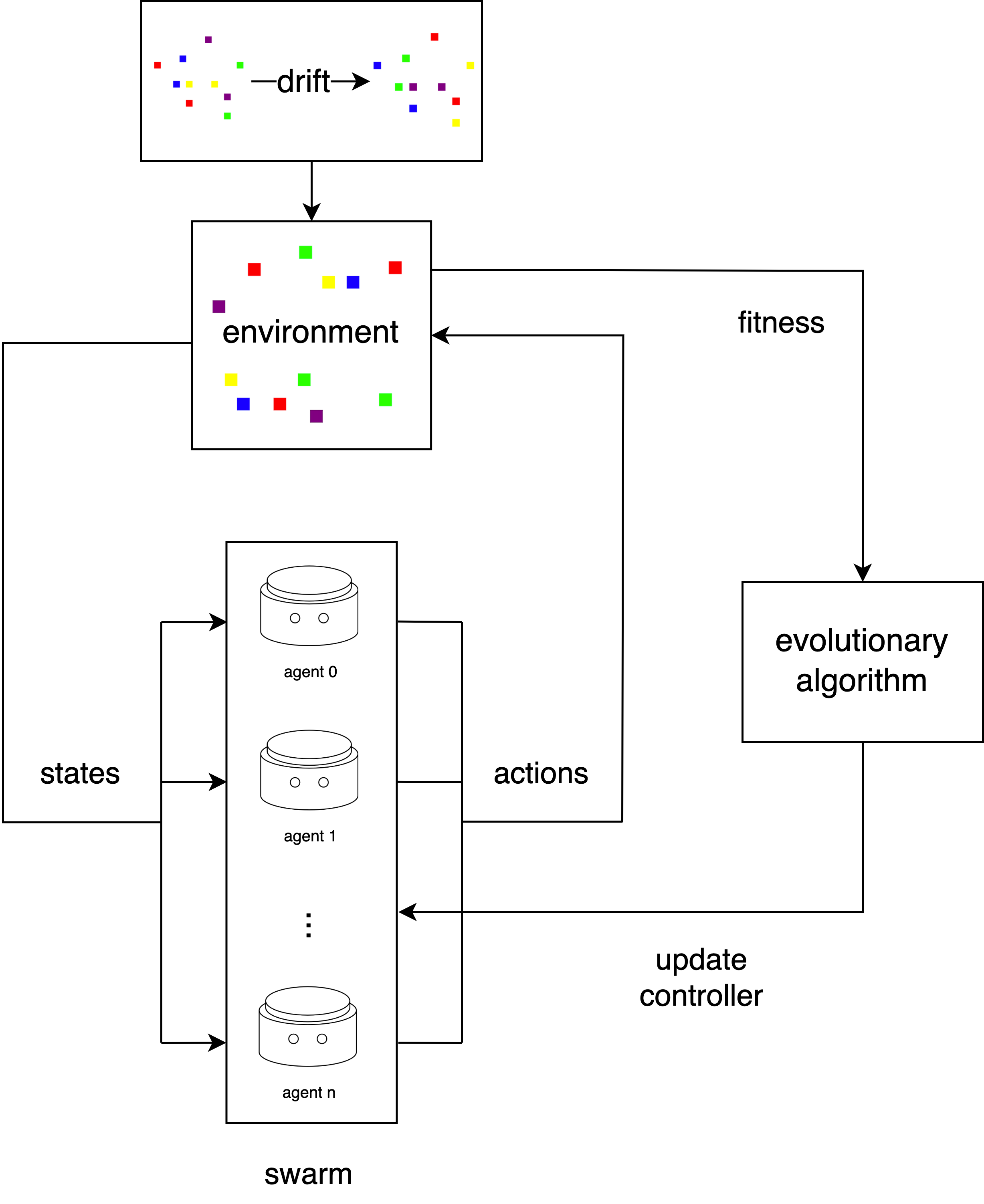}
    \caption{Overview of the lifelong evolutionary swarms framework. The swarm, composed of multiple agents, interacts with the environment through actions based on local sensor inputs (environment state) processed by an identical internal controller. Each agent keeps a copy of the controller. The sequence of actions of an agent determines the fitness of the controller. The evolutionary algorithm updates the controller based on their fitness. Periodically, the dynamic environment changes the underlying task which requires the swarm to adapt to novel conditions without forgetting the previous knowledge.}
  \label{fig:framework}
\end{figure}

\paragraph{Lifelong neuroevolution}
In the context of lifelong evolution of neural networks, \cite{Ellefsen2015} investigates how evolving modular neural networks can mitigate catastrophic forgetting by reducing interference between tasks. The setup involves an abstract environment where organisms evolve to maximize fitness by learning to consume nutritious food and avoid poisonous food, with abrupt seasonal changes introducing new food sources. By encouraging modularity through a connection cost mechanism, networks evolve to separate functionality into distinct modules, enabling selective learning. This modularity improves performance by allowing agents to learn new skills faster while retaining old ones. The study highlights that modularity not only enhances learning dynamics but may also reflect an evolutionary advantage observed in natural animal brains to combat catastrophic forgetting.\\
The study in \cite{Kashtan2007} explored how modularly varying goals (MVG) can accelerate evolutionary processes in computer simulations. Leveraging on genetic algorithms, the authors investigated how populations of networks evolve under temporally changing goals, where each new goal shared subproblems with ones. The tasks involved combinatorial logic problems, such as solving specific input-output relationships, which grew in complexity with modular goal changes. Their results showed that MVG significantly reduced the number of generations required to achieve a given goal compared to fixed-goal evolution.\\
To the best of our knowledge, we are the first to focus on lifelong neuroevolution of swarm controllers.
\begin{table}[t]
    \centering
    \caption{The input and output nodes for the swarm's neural controller. $c$ is the number of possible colors available in the arena and $n$ is the number of neighbors perceived.}
    \label{tab:in_out}
    \begin{tabular}{l c c}
    \toprule
    \textbf{Input} & \textbf{Nodes} & \textbf{Value} \\ \hline
         neighbor type &  $(c + 3) \times n$ & \{\textit{none, wall, agent, $\text{color}_{1,\ldots,c}$}\}\\
         neighbor distance & $n$ & [0, 1] \\
         neighbor direction & $2 \times n$ & [0, 1] \\
         heading & $2$ & [0, 1] \\
         carrying box & $c + 1$ & \{\textit{none, color$_1$, ..., color$_c$}\} \\
         target color & $c$ & \{\textit{color$_1$, ..., color$_c$}\} \\
    \hline
    \textbf{Output} & \textbf{Nodes} & \textbf{Value} \\ \hline
        wheel velocities & 3 & [$-v_{max}$, $v_{max}$] \\
    \bottomrule
    \end{tabular}
\end{table}
\section{An Environment for the Lifelong Evolution of Swarms} \label{sec:environment}
To study and address the challenges of i) quick adaptation in the presence of task drifts and ii) retention of previous knowledge in swarm controllers, we present our lifelong evolutionary framework (Figure \ref{fig:framework}). This section only assumes that the swarm controller is optimized with an evolutionary algorithm that keeps a population $P$ of controllers (individuals) $\pi \in P$. We defer the discussion about the evolutionary algorithm we adopted to Section \ref{sec:controller}.\\
Each agent in the swarm perceives the environment through their sensors. The perceived state is local to the agent. The same controller is deployed across all agents, forming a homogeneous swarm. It processes the local environment state, along with any additional external information, to determine each agent's next action. We define an \emph{episode-based} environment, where at the start of each episode the agents are positioned in the environment and they move and operate for a fixed amount of $K$ steps, at the end of which the episode terminates. A step involves each robot sensing the environment, providing the observations as input to the controller, generating output commands, sending them to the actuators, and moving to a new position. The swarm's objective is to solve a task, implicitly defined by a reward function $\mathcal{R}: \mathcal{S} \times \mathcal{A} \rightarrow \mathbb{R}$, which takes an environment state $\mathbf{s} \in \mathcal{S}$, an action $\mathbf{a} \in \mathcal{A}$ and returns a scalar reward. The rewards are summed across all $K$ steps. Notice that the proposed setup is not tied to a specific task, but it can be used with any reasonable reward function.\\
At each generation of the evolutionary algorithm, each controller is evaluated across $N$ randomly generated environments with varying initial states, but with the same reward function. The final fitness score is computed as the average total reward over the evaluation environments. Formally, the fitness for task $t$ of a controller $\pi$\footnote{Note that $\pi$ defines the controller phenotype (a function mapping states to actions). The evolutionary algorithm searches for fit phenotypes by acting on the corresponding genotypes.} is:
\begin{equation}
    f_t(\pi) = \frac{1}{N} \sum_{i=1}^{N} \sum_{k=1}^{K} \mathcal{R}_t(\mathbf{a}_k = \pi(\mathbf{s}_k), \mathbf{s}_k) .
\end{equation}
The controller receives the current state $s_k$ and returns the action $a_k$. The reward function, as the fitness, explicitly depends on the task $t$ (Section \ref{sec:setup} provides our choice for the reward function). The fitness is used by the evolutionary algorithm to update the controllers.\\
The key innovation of our framework is the introduction of dynamic target objectives that change (drift) over the population's lifetime. This sudden change mimics real-world environmental shifts where the relationship between actions and rewards is redefined. This corresponds to a change in the mapping implemented by $\mathcal{R}$.
As a result, the top-performing controller will see a decrease of its fitness, as it faces a new objective. We monitor the performance on previous tasks by adopting the same evaluation protocol mentioned above. For example, when evolving a controller on the second task, we measure its average fitness over $N$ randomly-generated environment according to the reward function of the second task (\emph{adaptation}), and on another $N$ environments according to the reward function of the first task (\emph{retention}). 

We formally define our evaluation metrics. We denote the retention of a (previous) task $t-1$ for a controller $ \pi $ in a population $ P $ as $f_{t-1}(\pi)$. Among all individuals in a population $P_t$ evolved for the \emph{current} task $t$, we are particularly interested in the one with the highest current (C) fitness:
\begin{equation}
    C(P_t) = \max_{\pi \in P_t} f_t(\pi) .
\end{equation}
This metrics quantifies the effectiveness of the evolutionary algorithm to find good controllers for the current task $t$.\\
We can consider retention at two levels:
\begin{itemize}
    \item \emph{Population Level}: we examine whether some individuals in the population retain knowledge from previous tasks. Such individuals are crucial as they can guide the evolutionary process to quickly re-adapt when re-encountering a previously faced task. A diverse population is expected to exhibit varying levels of retention across individuals. To capture this, we evaluate each individual on the previous task and identify the one that retains the highest performance ($R^{pop}$), i.e., the individual with the largest retention:
        \begin{equation}
        R^{pop}_{t-1}(P_t) = \max_{\pi \in P_t} f_{t-1}(\pi).
        \label{eq:ret_pop}
    \end{equation}
    This metric is aligned with a multi-agent system, where we always consider that there is a population of solutions, and not a single agent to be optimized.
    \item \emph{Individual Level}: we measure the retention ($R^{top}$) of the best individual for the current task on a previous task. This is particularly relevant when the population can no longer evolve, such as during deployment, where a single controller is needed for inference. This is also a more challenging objective, because the best-performing individual for the current task may evolve specific skills that are useful for that task, and less useful for others:
        \begin{equation}
        R^{top}_{t-1}(P_t) = f_{t-1} \left( \arg \max_{\pi \in P_t} f_t (\pi) \right) .
        \label{eq:ret_top}
    \end{equation}
    This view is aligned with the lifelong learning approach, where a single agent is optimized on a stream of tasks.
\end{itemize}
The main reason we are interested in assessing the retention of previous knowledge is that previous tasks may reappear in the future.  This is a well-known idea in lifelong learning, where an environment with repeating tasks can be exploited by the agent to improve its performance \cite{hemati2023}. Interestingly, a large body of works in lifelong learning focuses on environments that never repeat previous tasks, although the agent is still required to retain knowledge about them; a case which is not that common in the real-world \cite{cossu2022}.\\
We define \emph{forgetting} $F_{t-1}$ for task $t-1$ as the difference between the best performance previously achieved on that task and the retention after evolving for the subsequent task. Formally:
\begin{equation}
    F^{pop/top}_{t-1} = C(P_{t-1}) - R^{pop/top}_{t-1}(P_t) .
\end{equation}
Note that forgetting compares the performance on the \emph{same task} (same reward function) of \emph{two different populations}. Although forgetting can be computed for the population at any generation, it is most insightful when evaluated at the last generation of each task. \\

To experiment with our lifelong evolutionary framework, we chose \emph{foraging} tasks as a well-established benchmark in swarm robotics. Foraging requires the swarm to explore the environment, identify target resources and transport them to a pre-defined drop zone. Foraging is a useful benchmark for lifelong learning, as it can easily be extended to include changes in the reward function. Our implementation requires the swarm to locate a set of boxes marked by a target color. After a certain time, the target color changes and the swarm has to fetch the new boxes instead.


\subsection{Environment Setup} \label{sec:setup}
We implemented the foraging environment as a 2D continuous-space simulation using the Gymnasium\footnote{\url{https://gymnasium.farama.org}} library in Python \cite{towers2024}. The environment consists of a 5-meter $\times$ 5-meter arena populated with swarm of agents and a set of color-coded boxes scattered across the entire environment. The drop zone is positioned along the arena's upper edge. Once an agent drops a box, it goes on to fetch another one. The simulation expires after a fixed amount of time.\\
Each robot and box is represented as a point in the arena, with a diameter of 250 mm. Robots pick up boxes by coming into contact with them, defined as the robot’s center being within a 125 mm radius of the box’s center. Similarly, they drop boxes when positioned near the drop zone. Robots are equipped with omnidirectional wheels, enabling movement in any direction at a maximum speed of 50 cm/s.\\
Robots operate solely on individual sensor input to replicate realistic swarm scenarios. Robots lack knowledge of their global position, the location of boxes, or the drop zone.\\
To emulate the capabilities of real-world DOTS robots \cite{Jones2022}, which serve as the basis for our study, it is assumed that each robot can perceive the type, distance, and orientation of entities (including other robots, walls, and boxes) within a 1m range. Robots can also identify whether they are carrying a box, and determine their heading direction using an onboard compass. The target color can be provided as input to the controller if needed.\\
For the experiments, the arena is populated with a swarm of five robots and 20 boxes of various colors, with half matching the target color and the other half having a different one. Each simulation episode inlcudes $K=500$ steps, with each step representing 0.1 seconds. We used $N=10$ evaluation environments.\\
Positive individual behaviors, such as picking up a correctly colored box, earn one point, while delivering it to the drop zone earns two points. Conversely, picking up a wrongly colored box results in a penalty of -1 point. Given $\mathbf{s}, \mathbf{a}$ as two vectors defining the perceived state and resulting action of each agent in the swarm, we define our reward function for task $t$ as:
\begin{equation*}
    \mathcal{R}_t(\mathbf{s}, \mathbf{a}) = \sum_{(s_i, a_i) \in (\mathbf{s}, \mathbf{a})}
\begin{array}{l}
\begin{cases}
+1, & \text{picking up a target $t$ box,} \\
+2, & \text{delivering a target $t$ box,} \\
-1, & \text{picking up a non-target $t$ box,}\\
0 & \text{else.}
\end{cases}
\end{array}
\end{equation*}

\section{Lifelong Neuroevolution of Swarm Controllers} \label{sec:controller}
The swarm’s behavior is governed by an artificial neural network controller evolved using NEAT \cite{Stanley2002}. NEAT evolves both the network's topology and its weights. Network inputs consist of the robot’s local environmental observations, while the outputs correspond to velocity commands for the robot’s three wheels. Before being processed by the neural controller, observations are preprocessed. Categorical variables, such as the type of perceived entities or the type of box being carried, are one-hot encoded. Continuous variables are normalized to a range of 0 to 1, and angular directions are transformed into sine and cosine components, ensuring smooth representation and continuity between angles like 1° and 360°. A detailed breakdown of inputs and outputs is provided in \autoref{tab:in_out}.
\begin{figure}[t]
    \centering
    \includegraphics[width=0.8\linewidth]{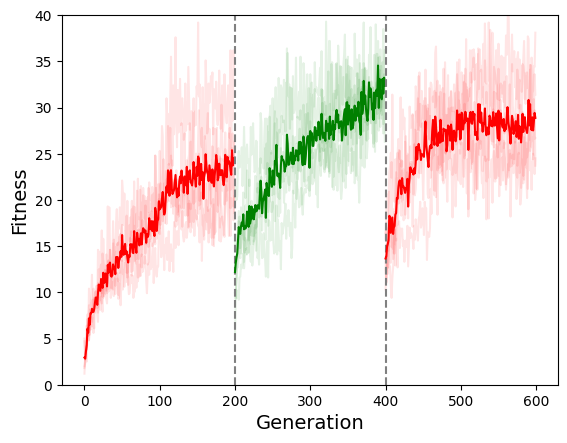}
    \caption{Fitness evolution of the best individual in the population for each generation and across three tasks: red task (generations 0–200), blue task (generations 200–400), and a return to the red task (generations 400–600). Line colors correspond to the task, and vertical dashed lines mark the transitions between tasks. The rapid recovery of fitness after task switches highlights the system's adaptability and transfer of knowledge between related tasks.}
    \label{fig:lifelong_evo}
\end{figure}

\paragraph{Mitigating forgetting via genetic distance regularization} As we will see in Section \ref{sec:forgettingind}, the best performing individual on a given task is subject to forgetting of previous knowledge. To mitigate this phenomenon we introduce a \textit{genetic distance} regularization method, inspired by Elastic Weight Consolidation (EWC) \cite{kirkpatrick2017} from lifelong learning. Our approach adds a penalty term to the fitness function, like EWC adds a penalty term to the loss function. EWC mitigates forgetting by pulling \emph{important} weights of the current network towards reference values that were effective for previous tasks. While originally designed for deep learning, we adapted the same principle for the evolutionary algorithm used in our framework by i) selecting important connections, ii) choosing a reference model, iii) pulling existing weights toward their reference. 

i) EWC uses the Fisher Information Matrix to select the important connections. Since we are dealing with a gradient-free evolutionary approach, the Fisher Information is not available. Our approach starts with a small network without hidden layers and evolve it using NEAT. This allows the architecture to expand dynamically and to add or remove connections and nodes. Therefore, we assume that all existing connections are important, as unnecessary ones would likely be removed during the evolutionary process.\\
ii) It is straightforward for EWC to choose the reference model, as it trains a single network. Evolutionary techniques like NEAT, instead, maintain a population of solutions and thus require a selection process to determine the best reference model (e.g., the simplest, the fittest, or the network which changed the least). We select the individual with the highest \emph{regularized fitness} (Equation \ref{eq:regf} below) as the reference model. Moreover, the network's architecture in NEAT changes across generations. Therefore, we need to map the architecture of the current individual to the reference network's architecture. 
\begin{figure}[t]
    \centering
    \includegraphics[width=0.8\linewidth]{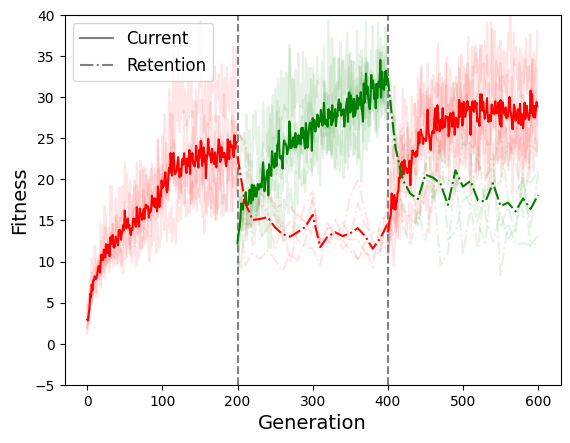}
    \caption{Current performance (solid lines) and retention at the population level (dashed lines) across task drift (line colors correspond to task colors). Population naturally preserves past knowledge while adapting to new objectives.}
    \label{fig:retention_pop}
\end{figure}
Given two controller's genotypes $x_1$ and $x_2$, we use the NEAT routine $\delta$ to compute their genetic distance:
\begin{equation}
    \delta(x_1, x_2) = \frac{c_1 E(x_1, x_2)}{\text{max}\{|x_1|,|x_2|\}} + \frac{c_2 D(x_1, x_2)}{\text{max}\{|x_1|,|x_2|\}} + c_3 \cdot \overline{W}(x_1, x_2).
\end{equation}
The function is a linear combination of the number of excess genes $E$ (present after the last matching gene), disjoint genes $D$ (non-matching genes between aligned matching genes), and the average weight difference of matching genes $\overline{W}$ (including disabled ones), weighted by coefficients $c_1$, $c_2$, $c_3$ and normalized based on the largest genome size.\\
iii) Our approach computes the \emph{regularised fitness} as:
\begin{equation} \label{eq:regf}
    f^{gd}_t(\pi) = f_t(\pi) - \lambda \delta(x_{\pi^*}, x_{\pi}),
\end{equation}
$x_{\pi}$ denotes the genotype of an individual $\pi$ from the current population, and $x_{\pi^*}$ is the genotype of the reference model $\pi^*$ from the previous task. The term $\lambda$ is a weighting coefficient that controls the influence of the genetic distance penalty. \\
As with other regularization techniques in lifelong learning, we expect a trade-off between the performance on the current task and the performance on previous tasks. A high regularization will prevent the population to adapt, while a low regularization will not preserve previous knowledge.

\section{Experiments}
We evaluate our lifelong evolutionary framework\footnote{Full implementation available at \url{https://github.com/lorenzoleuzzi1/lifelong_evolutionary_swarms}.} for swarms according to i) their ability to adapt to novel tasks, including transferring knowledge from previous tasks to increase the performance on the current one; ii) their ability to retain knowledge from previous tasks; iii) the ability of the top-performing individual to mitigate forgetting with our genetic distance regularization.\\
We present results averaged over five random seeds that control the initial network configuration, the stochastic genetic operations (recombination and mutation), and the initial agents and boxes location. This means that runs that achieve a similar fitness values may end up with very different networks, potentially belonging to different NEAT species. The complete NEAT configuration can be found in Appendix \ref{app:neat}.

\begin{figure}[t]
    \centering
    \includegraphics[width=0.8\linewidth]{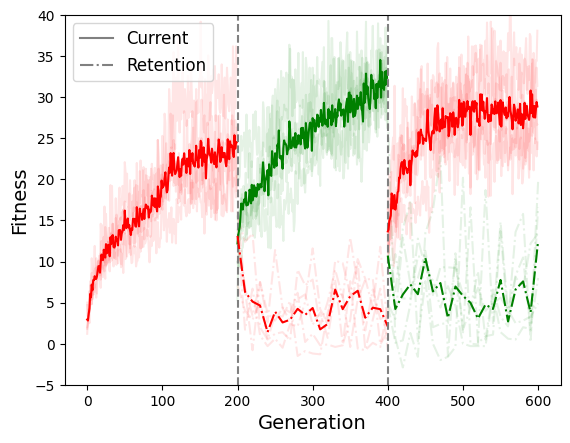}
    \caption{Current performance (solid lines) and retention at the individual level (dashed lines) across task drift (line colors correspond to task colors). Retention fitness demonstrates catastrophic forgetting, as the best-performing individual on the current task fails to retain knowledge of prior tasks.}
    \label{fig:retention_top}
\end{figure}

\subsection{Quick Adaptability} \label{sec:quickadapt}
We consider a dynamic environment where the task switches from retrieving red boxes to retrieving green boxes and then it switches back to the red boxes. Upon task change, the swarm should be able to quickly adapt to the new task (after a short transient). We also expect the swarm to reuse previous knowlegde. This means that, whenever a previous task reappears, the swarm's performance should be higher than the one of a randomly-initialized swarm controller.\\
\autoref{fig:lifelong_evo} shows the fitness of the top individual for each generation across the three tasks. We observe that at generation 0 the swarm is unable to retrieve boxes (fitness around 0), as the controller is still a randomly initialized one. As the evolution progresses the fitness steadily increases up until 25 at generation 200, when the task drifts to the green boxes. The fitness then suddenly decreases as the previously evolved controller still retrieves red boxes. However, it does not drop to 0, highlighting that the swarm can indeed exploit knowledge from the previous task. Morever, the swarm adapts much quicker than before, reaching the same fitness of the red task in less than half of the generations previously required. This is further evidence that prior learning on the red task has facilitated faster adaptation to the new objective. \\
The same rapid recovery is also observed after generation 400, when the red task reoccurs. Within the first 50 generations, the previous performance on the red task achieved at generation 200 is fully restored. The lifelong evolutionary process is able to leverage prior knowledge in order to speed up convergence on new tasks, requiring significantly fewer generations. Compared to a static evolution setup, where each task is evolved separately and by starting from a randomly-initialized population, lifelong evolution requires less resources and can effectively transfer knowledge across similar tasks.

For a better understanding of how fitness levels correspond to actual performance, such as the number of boxes retrieved, refer to the episode visualizations provided in the supplementary material (Appendix \ref{sec:materials}).

\begin{figure}[t]
    \centering
    \includegraphics[width=0.8\linewidth]{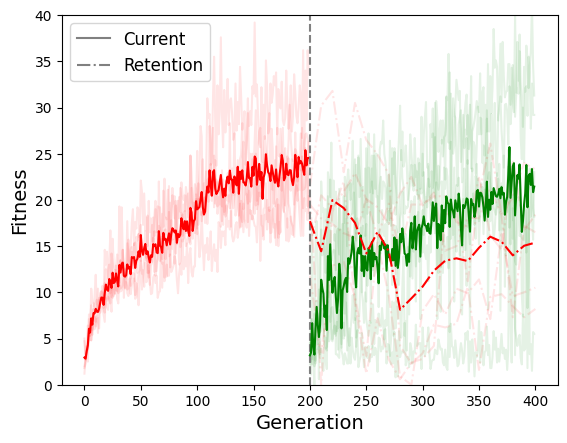}
    \caption{Current performance (solid lines) and retention at the individual level (dashed lines) for the red and green tasks when applying a fixed regularization coefficient,  $\lambda = 11$. Retention is higher than without regularization (mitigating forgetting), though at the cost of a reduced performance on the green task.}
    \label{fig:ret_11}
\end{figure}
\begin{figure}[t]
    \centering
    \includegraphics[width=0.8\linewidth]{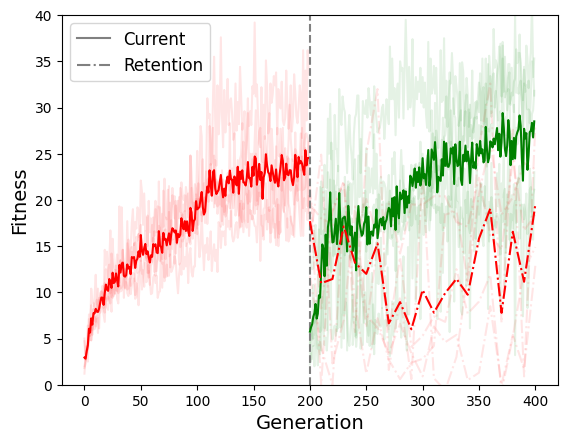}
    \caption{Current performance (solid lines) and retention at the individual level (dashed lines) for the red and green tasks when using model-specific regularization coefficients. This setup improves retention and the performance on the green task compared to \autoref{fig:ret_11}.}
    \label{fig:ret_diff}
\end{figure}

\subsection{Forgetting in the population} \label{sec:forgettingpop}
In the spirit of lifelong learning, we measure the ability of the evolved population to mitigate forgetting by retaining previous knowledge. We use the same dynamic environment of Section \ref{sec:quickadapt}. Every 10 generations we evaluate the fitness on the previous task of each individual in the population. \autoref{fig:retention_pop} shows the fitness of the best-performing individual at each generation according to \autoref{eq:ret_pop}. We explicitly plot the retention curve, while the forgetting is obtained by comparing the dashed red curve at generation 400 (performance on previous red task when evolving on the green task) with the solid red curve at generation 200 (performance on the red task at the end of evolution on the red task).

The red retention curve shows that at each generation in the green task there is always an individual in the population able to preserve some knowledge about the red task. The retention never drops below 10. The forgetting for the red task at the end of the evolution on the green task is 9.9 fitness points. We observe a similar behavior when switching back to the red task and measuring the retention on the green task (from generation 400). The forgetting in this case is 14.2 fitness point (with a final performance on the green task of around 32 points). Overall, we can see that lifelong evolution did not prevent the swarm adapting to a novel task \emph{and} it allowed individuals in the population to preserve knowledge about prior tasks \emph{without being explicitly evolved for this purpose.} Compared to the forgetting typically exhibited by lifelong learning agents in deep learning \cite{masana2023}, where the network completely forgets previous task if no specific lifelong techniques are used, the lifelong evolution of swarms is much more robust to forgetting.
This is likely due to the variety of the population maintained by evolutionary algorithms, like NEAT, that preserve previously discovered skills even though they may not be immediately useful \cite{mouret2015, pugh2016}.
\begin{table}[t]
    \centering
    \caption{Summary of results. "Red" and "Green" represent the current performance on their respective tasks. "Ret. Red" indicates the retained performance on the red task after evolving on the green task. Conversely, "Fgt. Red" measures forgetting on the red task after evolving on the green task. The approaches include population-based evolution (\emph{pop}), top-performing individual (\emph{ind}), regularized evolution with a fixed regularization coefficient (\emph{reg}) and with a model-specific coefficient (\emph{m.s. reg}).}
    \label{tab:summary}
    \begin{tabular}{ccccc}
    \toprule
         & \textbf{Red $\uparrow$} & \textbf{Green $\uparrow$} & \textbf{Ret. Red $\uparrow$} & \textbf{Fgt. Red $\downarrow$} \\
         \hline
         \emph{pop} & 24.54 & 32.36 & 14.64 & 9.9 \\
         \emph{ind} & 24.54 & 32.36 & 2.08 & 22.46 \\
         \emph{reg} & 24.54 &  21.44 & 15.38 & 9.16 \\
         \emph{m.s. reg} & 24.54 & 28.5 & 19.52 & 5.02 \\
    \bottomrule
    \end{tabular}
\end{table}
\subsection{Forgetting in the individuals} \label{sec:forgettingind}
Lifelong deep learning develops a single agent that needs to be capable to solve all encountered tasks. We study our lifelong evolutionary framework also from this perspective, to understand the benefits and pitfalls of an agent-centric view. We adopt the same dynamic environment of the previous sections. We select the top-performing individual on the current task and we evaluate its performance on previous tasks. Note that this approach differs from Section \ref{sec:forgettingpop}, as there we selected the individual which performed best \emph{on the previous task}. The results, shown in \autoref{fig:retention_top}, are in stark contrast to the population-level retention. The top individual catastrophically forgets previous knowledge, with retention scores dropping to near-zero across all generations. Specifically, forgetting for the red task at generation 400 is 22.5, and for the green task at generation 600 is 19.9. Therefore, while lifelong evolution supports retention at the level of the whole population, it does not do the same for each individual. These results are aligned with the behavior commonly observed in deep lifelong learning \cite{masana2023}.\\
\begin{figure}[t]
    \centering
    \includegraphics[width=0.8\linewidth]{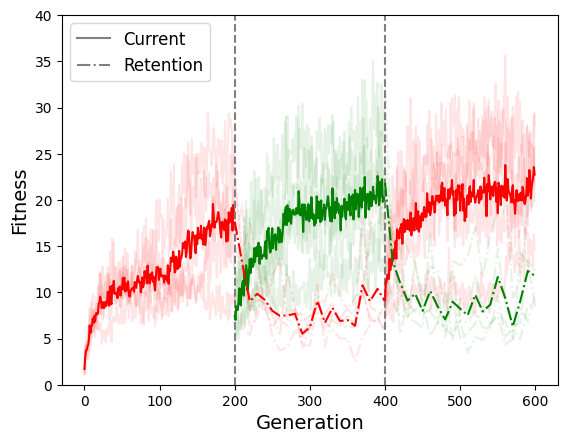}
    \includegraphics[width=0.8\linewidth]{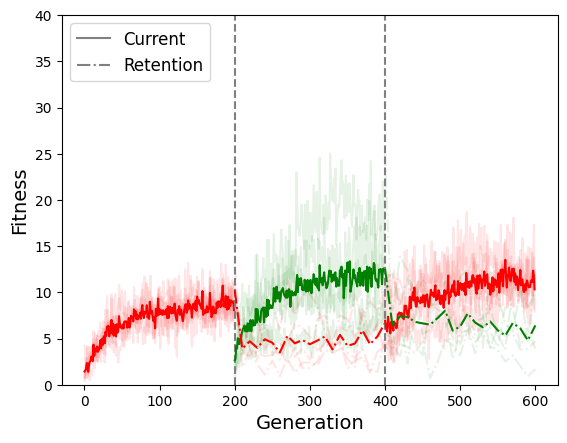}
    \caption{Impact of population scaling on current performance (solid lines) and retention at the population level (dashed lines). The top plot shows results for a population size of 100, while the bottom plot displays results for a reduced population size of 15. Although the population size significantly affects the overall fitness levels (higher fitness with a larger population), a smaller population is still able to preserve some knowledge about previous tasks.}
    \label{fig:scaling_pop}
\end{figure}
We tweak the evolution process with our genetic regularization, choosing as the reference model the best performing individual on the red task right before the switch to the green task (generation 200). 
\autoref{fig:ret_11} shows the results of applying the same regularization coefficient $\lambda=11$ on all runs. The retention curve is notably higher compared to the one without regularization (\autoref{fig:retention_top}), reaching 15.38 at generation 400 and reducing forgetting to 9.16. In contrast, the prior experiment exhibited a complete performance drop with a forgetting of 22.5.\\
Aligned with the deep lifelong learning literature \cite{parisi2019, kirkpatrick2017}, the regularization comes with a trade-off: the performance on the green task decreased from 32.36 (without regularization) to 21.44.

\paragraph{Model-specific regularization} So far, we used the same $\lambda$ coefficient for all runs. However, the coefficient directly impacts on architectural changes and, as such, different architectures may need different values of $\lambda$. To test this hypothesis, we tuned $\lambda$ on each run separately (each run using a different random seed). Aligned with our expectations, \autoref{fig:ret_diff} shows that we can indeed improve both retention and the performance on the current task with a model-specific $\lambda$. The retention for the red task at generation 400 reaches 19.52, reducing forgetting to just 5.02.\\
We select the top-performing controller from the model-specific regularization runs and evaluate it across 100 environments per task. It achieves a fitness score of 28.5 on the green task and 26.1 on the red task, with a forgetting score of only 2.2, demonstrating remarkable multitask learning capabilities \cite{caruana1997}.

\paragraph{Summary} \autoref{tab:summary} summarizes our results. The model-specific regularization retained the most knowledge at the expenses of a slight performance drop on the green task. Even with a fixed $\lambda$ value, a favorable balance between retention and current performance was achieved. At the population level, the controller is able to adapt to new tasks and to preserve some knowledge about previous ones.
\begin{figure}[t]
    \centering
    \includegraphics[width=0.8\linewidth]{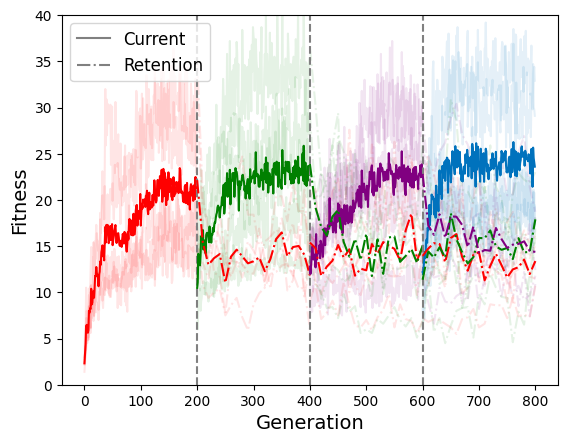}
    \caption{Current performance (solid lines) and retention at the population level (dashed lines) across four sequential tasks (red, green, purple, and blue). The population maintains sufficient diversity to retain knowledge on all previous tasks, while adapting to new ones.}
    \label{fig:more_drifts}
\end{figure}

\subsection{Stress tests on the population}
We turn our attention to two key factors of our lifelong evolutionary process: the population size and the number of task drifts.

\paragraph{Reducing the population size.} We study whether smaller populations are still able to preserve previous knowledge. Starting from the same dynamic environment of Section \ref{sec:forgettingpop}, we reduce the population from 300 to 100 individuals (\autoref{fig:scaling_pop}, top) and as low as 15 individuals (\autoref{fig:scaling_pop}, bottom). For the former case, forgetting is $9.38$, similar to what we achieved for a population of $300$. For the latter, forgetting is $1.38$ but only because the performance on the first task is very low. Therefore, smaller populations forget less only because they are unable to adapt properly. When the population size enables effective adaptation, forgetting remains comparable.  
Appendix \ref{app:morepop} shows the performance for an increased population of 600 individuals, which confirms the results reported here.  
\begin{figure}[t]
    \centering
    \includegraphics[width=0.8\linewidth]{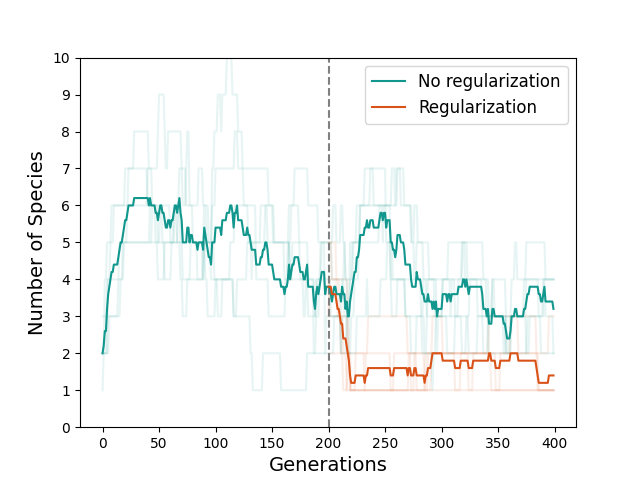}
    \caption{Number of species over generations with and without regularized evolution.}
    \label{fig:n_species}
\end{figure}
\paragraph{Increasing the number of drifts.} Using a population of 300 individuals, we increase the number of drifts to include four different tasks: red, green, purple and cyan.
Since we need to distinguish between more colors (eight in total, two per task), the input neurons for the neural networks are increased correspondigly (\autoref{tab:in_out}).\\
We evaluate the performance on \emph{all} previous tasks at the population level. \autoref{fig:more_drifts} shows that the population successfully retains previous knowledge at a similar score, without being significantly impacted by the addition of new tasks.

\paragraph{Variation of species.} We  analyze the number of NEAT species present in the population at each generation to understand how its diversity changes over time (\autoref{fig:n_species}).
When evolving without regularization the evolution starts with around two species, which increase to 6 by generation 50. Over time, species with stagnating performance are gradually discarded. This behavior repeats when the task changes.
In contrast, the regularized evolution shows a significant reduction in species diversity after the task change, when the genetic distance penalization is applied. The population often converges to one or very few species, which aligns with the objective of optimizing individuals to remain similar to the single reference model. While this approach enhances retention and mitigates forgetting, it reduces diversity. Incorporating mechanisms to preserve diversity around the reference model may improve evolvability \cite{mengistu2016, lehman2011b} and sustain evolution over longer time spans.

\section{Conclusion and Future Work}
We introduced the lifelong evolutionary framework for swarms and we implemented one instance based on a dynamic foraging task where the swarm controller is evolved via NEAT. We showed that the population of controllers is able to adapt to novel tasks, and to retain part of previous knowledge for future reuse: this includes quick adaptation when encountering previous tasks and forward transfer to new tasks. We also proposed a genetic regularization technique to increase retention at the level of the top-performing individual on the current task, which would otherwise catastrophically forgets previous knowledge.\\

We believe our empirical results provide solid foundations for future work. Other lifelong techniques may be adapted into our framework (e.g., replay \cite{hayes2021, merlin2022}) to improve retention. Lifelong learning may also benefit from the collective approach of evolutionary algorithms, with opportunities to naturally mitigate forgetting by maintaining an archive of previously useful skills \cite{lehman2011b} in the form of compressed models. Lifelong evolutionary approaches can have an impact well beyond lifelong learning and swarm intelligence alone. Letting the evolution and learning processes co-design an adaptive system is indeed one of the most promising and interesting research directions we can envision for the future.

\bibliographystyle{plain}
\bibliography{les}
\newpage
\ 
\newpage
\appendix
\section{Additional experiments}
\subsection{Larger population size} \label{app:morepop}
\begin{figure}[t]
    \centering
    \includegraphics[width=0.8\linewidth]{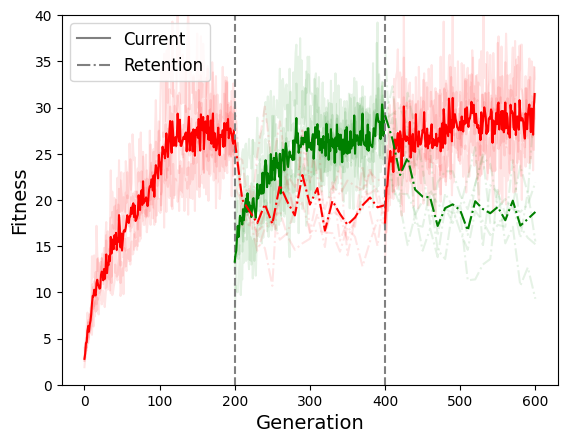}
    \caption{Current performance (solid lines) and retention at the population level(dashed lines) across task drift (line colors correspond to task colors). Population size is 600.}
    \label{fig:scaling_pop_big}
\end{figure}

\autoref{fig:scaling_pop_big} demonstrates that increasing the population size leads to improved fitness levels while maintaining the same lifelong learning dynamics, with red task forgetting of 8.06.
\subsection{Stronger regularization}
\begin{figure}[t]
    \centering
    \includegraphics[width=0.8\linewidth]{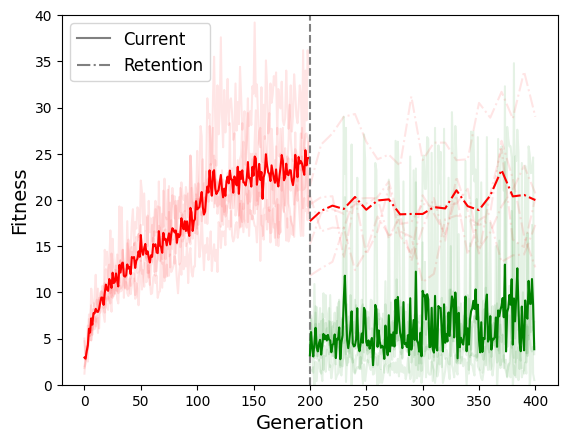}
    \caption{Current performance (solid lines) and retention at the individual level (dashed lines) for the red and green tasks when applying a strong regularization coefficient,  $\lambda = 20$.}
    \label{fig:reg_20}
\end{figure}

\autoref{fig:reg_20} illustrates the effect of a high regularization coefficient, where the evolution fails to adapt to the new task. The strong constraint to remain close to the previous reference model inhibits learning and prevents effective adaptation.
\subsection{Retention at the top-individual level for more drifts}
\begin{figure}[t]
    \centering
    \includegraphics[width=0.8\linewidth]{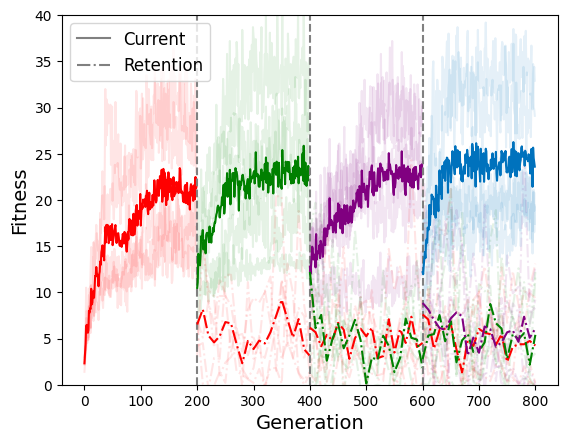}
    \caption{Current performance (solid lines) and retention at the individual level (dashed lines) across four sequential tasks (red, green, purple, and blue). The current top performing individual tends to forget previous tasks.}
    \label{fig:more_drifts_top}
\end{figure}

\autoref{fig:more_drifts_top} illustrates the retention of knowledge at the top-individual level across experiments involving four task drifts: red, green, purple, and cyan. The results highlight catastrophic forgetting, with retention curves being substantially lower than those observed when considering retention at the population level (\autoref{fig:more_drifts}).
\section{Species tracking}
\begin{figure}[t]
    \centering
    \includegraphics[width=0.8\linewidth]{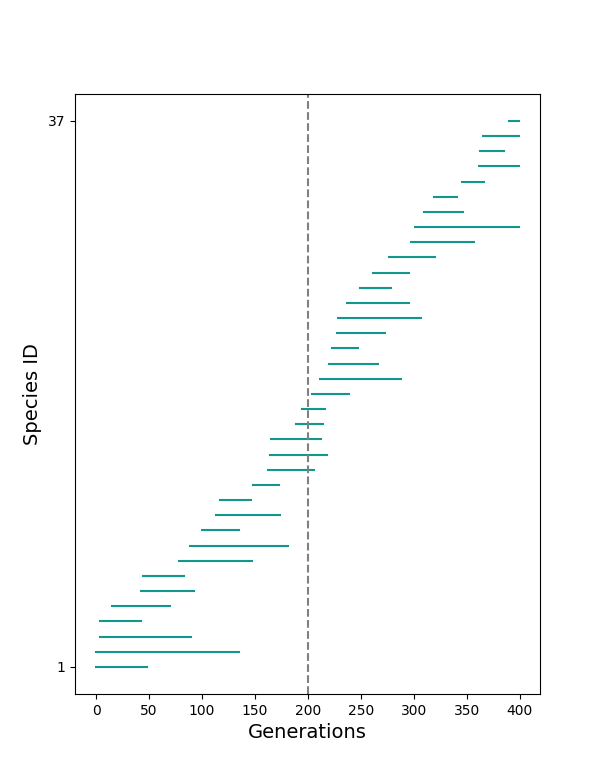}
    \includegraphics[width=0.8\linewidth]{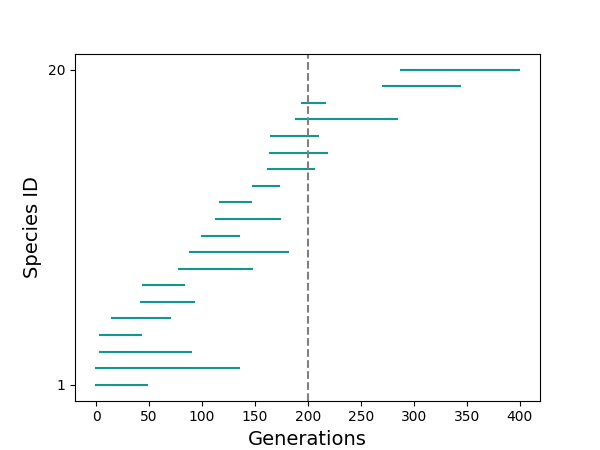}
    \caption{Lifespan of each species in evolution without regularization (top) and in evolution with regularization (bottom).}
    \label{fig:tracking_species}
\end{figure}

To visualize the impact of regularization on NEAT's speciation mechanism, \autoref{fig:tracking_species} tracks the lifespan of each species, showing when they emerge and when they go extinct. With regularization (bottom), after the first task (200 generations) there are significantly fewer species, and they tend to persist longer. In contrast, the standard non-regularized evolution (top) exhibits a more dynamic speciation process, with species constantly appearing and disappearing.

\section{NEAT configuration} \label{app:neat}
We use a population of 300 individuals and evolve the population for 200 generations on each task. NEAT starts with one hidden neuron and 50\% of active connections, initialized randomly from a standard Gaussian distribution, and clamped between -5 and 5. NEAT adds or deletes a node with 20\% probability, and mutates a weight by adding a random value sampled from a standard Gaussian with 80\% probability.\\
We utilized the \texttt{neat-python} library \cite{neat-python}, which implements NEAT. The full configuration details are provided in \autoref{tab:neat_config}. Notably, the compatibility threshold, which determines whether individuals belong to the same species based on genomic distance, is set to 3. Individuals with a genomic distance below this threshold are classified within the same species. We employed a modified sigmoidal transfer function, $\phi(x) = \frac{1}{1 + e^{-4.9x}},$ as suggested in \cite{Stanley2002}. This steepened sigmoid improves precision at extreme activations and is nearly linear in the range between -0.5 and 0.5, optimizing its steepest ascent. \\
For a detailed description of all other configuration parameters, refer to the official documentation\footnote{\url{https://neat-python.readthedocs.io/en/latest/config_file.html}}.
\begin{table}
\centering
\caption{NEAT Configuration Parameters}
\begin{tabular}{ll}
\toprule
\textbf{Parameter}                  & \textbf{Value}           \\
\midrule
num\_hidden                        & 1                       \\
initial\_connection                & partial\_direct 0.5     \\
feed\_forward                      & True                    \\
compatibility\_disjoint\_coefficient & 1.0                   \\
compatibility\_weight\_coefficient  & 0.6                    \\
conn\_add\_prob                    & 0.2                     \\
conn\_delete\_prob                 & 0.2                     \\
node\_add\_prob                    & 0.2                     \\
node\_delete\_prob                 & 0.2                     \\
activation\_default                & neat\_sigmoid           \\
activation\_options                & neat\_sigmoid           \\
activation\_mutate\_rate           & 0.0                     \\
bias\_init\_mean                   & 0.0                     \\
bias\_init\_stdev                  & 1.0                     \\
bias\_replace\_rate                & 0.1                     \\
bias\_mutate\_rate                 & 0.7                     \\
bias\_mutate\_power                & 0.5                     \\
bias\_max\_value                   & 5.0                     \\
bias\_min\_value                   & -5.0                    \\
response\_init\_mean               & 1.0                     \\
response\_init\_stdev              & 0.0                     \\
response\_replace\_rate            & 0.0                     \\
response\_mutate\_rate             & 0.0                     \\
response\_mutate\_power            & 0.0                     \\
response\_max\_value               & 5.0                     \\
response\_min\_value               & -5.0                    \\
weight\_max\_value                 & 5                       \\
weight\_min\_value                 & -5                      \\
weight\_init\_mean                 & 0.0                     \\
weight\_init\_stdev                & 1.0                     \\
weight\_mutate\_rate               & 0.8                     \\
weight\_replace\_rate              & 0.1                     \\
weight\_mutate\_power              & 1.0                     \\
enabled\_default                   & True                    \\
enabled\_mutate\_rate              & 0.01                    \\
compatibility\_threshold           & 3.0                     \\
species\_fitness\_func             & max                     \\
max\_stagnation                    & 20                      \\
species\_elitism                   & 1                       \\
elitism                            & 5                       \\
survival\_threshold                & 0.2                     \\
\bottomrule
\end{tabular}
\label{tab:neat_config}
\end{table}

\section{Regularization coefficient search details}
To determine the optimal regularization coefficient ($ \lambda $), a systematic search was conducted. Since fitness values typically ranged between 20 and 30, we used this as a reference for the magnitude of penalization. An initial coarse search was performed with $\lambda$ values of 5, 10, and 15. Based on the most promising results, we refined the search around 10 by experimenting with values 8, 9, 11, and 12.

The value $ \lambda = 11 $ was found to perform best across all runs in general. However, as noted, applying the optimal coefficient tailored to each run yields even better results. \autoref{tab:lambdas} presents the specific $ \lambda $ values associated with each run.
\begin{table}[]
    \centering
    \caption{Regularization coefficient $\lambda$ and seed associated to each run.}
    \label{tab:lambdas}
    \begin{tabular}{ccc}
    \toprule
        & seed & $\lambda$ \\ \hline
         run1 & 13 & 5 \\
         run2 & 17 & 5 \\
         run3 & 24 & 5 \\
         run4 & 31 & 11 \\
         run5 & 42 & 11 \\
    \bottomrule
    \end{tabular}  
\end{table}
\section{Supplementary Materials}
\label{sec:materials}
We includes GIFs showcasing the controllers in action during random episodes. Detailed instructions for viewing and interpreting the animations can be found in the accompanying \texttt{readme.txt}. Access the animations at: \url{https://drive.google.com/drive/folders/17fESIvcVOXVga5L8U-H5Vdv_zJ9bSYrB?usp=sharing}

We provide the resulting neural controller topologies from each evolution, categorized by tasks and different runs. Detailed explanations are provided in the \texttt{readme.txt}. Access the neural controllers at:  \\
\url{https://drive.google.com/drive/folders/1H3nhtlvedd-ILR7gcHMpR9E7g_QouSda?usp=sharing}
\section{Swarm foraging environment details}
The \texttt{SwarmForagingEnv}\footnote{\url{https://github.com/lorenzoleuzzi1/lifelong_evolutionary_swarms/blob/main/environment.py}} 
is a custom reinforcement learning environment designed for multi-agent swarm robotics tasks, where agents collaborate to retrieve specific target boxes in a dynamic and configurable environment. The environment supports episodic tasks and incorporates mechanisms for simulating task changes. The environment implements standard Gym API methods, including \texttt{step} and \texttt{reset}. Additionally, a \texttt{change\_task} method enables updating the target color and the set of present colors, limited to those defined during initialization. In our experiments, each task features 2 unique colors, ensuring no overlap with colors from other tasks. For example, during the red task, boxes are red and blue, while during the green task, boxes are green and yellow. \autoref{tab:swarm_foraging_env} details the initialization attributes of the class, including the values set for the experiments.
\begin{table}[ht]
    \centering
    \caption{\texttt{SwarmForagingEnv} class initialization parameters}
    \label{tab:swarm_foraging_env}
    \begin{tabular}{|l|l|p{0.35\linewidth}|}
        \hline
        \textbf{Parameter Name} & \textbf{Value} & \textbf{Description} \\ \hline
        \texttt{size} & 20 units & Size of the simulation arena, represented as a grid. \\ \hline
        \texttt{target\_color} & Varying & Target color for the boxes that agents must retrieve. \\ \hline
        \texttt{n\_agents} & 5 & Number of agents in the swarm. \\ \hline
        \texttt{n\_boxes} & 20 & Number of boxes in the arena. \\ \hline
        \texttt{n\_neighbors} & 3 & Number of nearest neighbors detected by an agent's sensors. \\ \hline
        \texttt{sensor\_range} & 4 units & Maximum sensing range of agents. \\ \hline
        \texttt{max\_wheel\_velocity} & 2 units/s & Maximum wheel velocity for agent movement. \\ \hline
        \texttt{sensitivity} & 0.5 units & Threshold distance for agents to interact with a block. \\ \hline
        \texttt{time\_step} & 0.1 s & Time step duration for each simulation step. \\ \hline
        \texttt{duration} & 500 & Maximum number of steps per episode. \\ \hline
        \texttt{max\_retrieves} & 20 & Maximum number of boxes that can be retrieved in one episode. \\ \hline
        \texttt{colors} & Varying & List of colors available for the boxes. \\ \hline
        \texttt{season\_colors} & Varying & List of colors available during a specific season. \\ \hline
        \texttt{rate\_target\_block} & 0.5 & Proportion of target boxes among all boxes in the arena. \\ \hline
        \texttt{repositioning} & \texttt{True} & Whether boxes are repositioned after being retrieved. \\ \hline
        \texttt{efficency\_reward} & \texttt{False} & Whether to reward agents for completing tasks before the maximum steps. \\ \hline
        \texttt{see\_other\_agents} & \texttt{False} & Whether agents can detect other agents in the arena. \\ \hline
        \texttt{boxes\_in\_line} & \texttt{False} & Whether boxes are placed in a line streight line during initialization. \\ \hline
    \end{tabular}
\end{table}
\end{document}